\documentclass[10pt,twocolumn,letterpaper]{article}

\usepackage{iccv}
\usepackage{times}
\usepackage{epsfig}
\usepackage{graphicx}
\usepackage{amsmath}
\usepackage{amssymb}

\graphicspath{{./graphics/}}
\usepackage{subcaption}
\usepackage{booktabs}
\usepackage{algorithm}
\usepackage[noend]{algpseudocode}
\usepackage{enumitem}

\usepackage[pagebackref=true,breaklinks=true,letterpaper=true,colorlinks,bookmarks=false]{hyperref}

\iccvfinalcopy %

\def\iccvPaperID{5748} %

\ificcvfinal\fi

\newcommand{\boldspacepar}[1]{\par\smallskip\noindent\textbf{#1}}

\begin{document}

\title{Rehearsal revealed: \\The limits and merits of revisiting samples in continual learning}

\author{Eli Verwimp\thanks{Authors contributed equally.}\\
KU Leuven\\
{\tt\small eli.verwimp@kuleuven.be}
\and
Matthias De Lange\footnotemark[1]\\
KU Leuven\\
{\tt\small matthias.delange@kuleuven.be}
\and
Tinne Tuytelaars\\
KU Leuven\\
{\tt\small tinne.tuytelaars@kuleuven.be}
}

\maketitle
\ificcvfinal\thispagestyle{empty}\fi

\begin{abstract}
Learning from non-stationary data streams and overcoming catastrophic forgetting still poses a serious challenge for machine learning research. Rather than aiming to improve state-of-the-art, in this work we provide insight into the limits and merits of rehearsal, one of continual learning's most established methods. We hypothesize that models trained sequentially with rehearsal tend to stay in the same low-loss region after a task has finished, but are at risk of overfitting on its sample memory, hence harming generalization. We provide both conceptual and strong empirical evidence on three benchmarks for both behaviors, bringing novel insights into the dynamics of rehearsal and continual learning in general. Finally, we interpret important continual learning works in the light of our findings, allowing for a deeper understanding of their successes.\footnote{
Code: {\scriptsize \url{https://github.com/Mattdl/RehearsalRevealed}}}   
\end{abstract}

\section{Introduction}
\label{sec:intro}

Recent advances of neural networks have shown promising results by surpassing human capabilities in a wide range of tasks \cite{silver2018general,russakovsky2015imagenet,silver2016mastering}.
However, these tasks are typically highly confined and remain static after deployment.
This stems from a major limitation in neural network optimization, namely the assumption of independent and identically distributed (iid) training and testing distributions.
When the iid assumption is not satisfied during learning, neural networks are prone to catastrophic forgetting~\cite{french1999catastrophic}, causing them to completely forget previously acquired knowledge. 

\begin{figure}[!ht]
    \centering
    \includegraphics[width=0.95\linewidth]{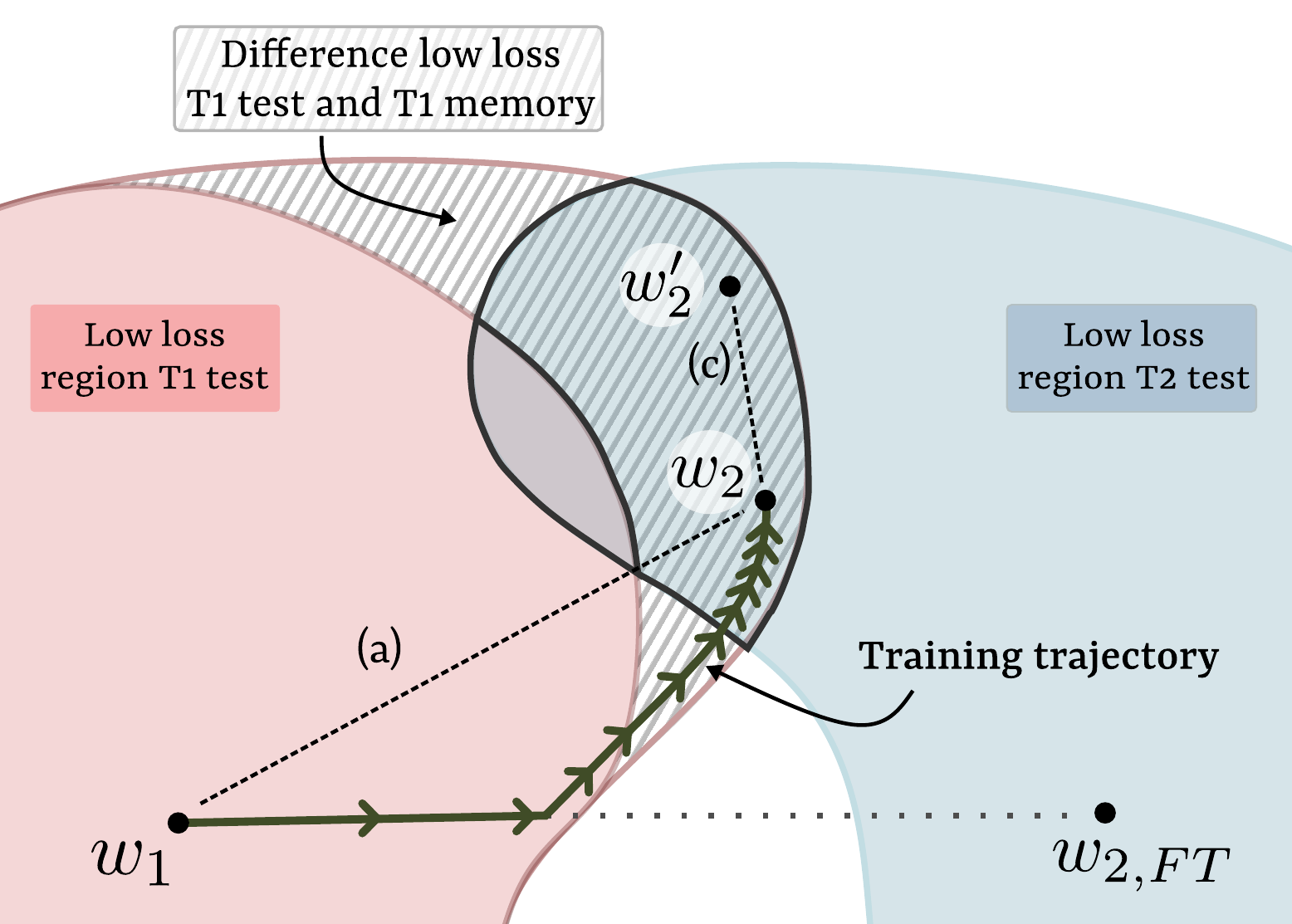}
    \caption{\label{fig:setup}Illustration of our findings, visualized as the loss values in parameter space. When using rehearsal to train task 2 ($T2$) after training task 1 ($T1$), i.e. from $w_1$ to $w_2$, the learning trajectory will first move towards a finetuned minimum $w_{2, FT}$ but deflects at a high-loss ridge of $T1$'s low-loss region. 
    The crux is $w_2$ ending up in the striped area where low loss for $T1$ memory is in contrast to the higher loss observed for $T1$ test data. 
    Additionally, we show empirically that linear low loss paths (a) and (c) exist between $w_1$ and $w_2$, and between two models trained with different rehearsal memories $w_2$ and $w_2'$. }
\end{figure}

Continual or lifelong learning strives to overcome this static nature of neural networks with a wide range of mechanisms~\cite{9349197}, among which \emph{rehearsal} has shown promising results~\cite{lopez2017gradient,Rebuffi2017,chaudhry2019continual,de2020continual}.
Rehearsal aims to approximate the observed input distributions over time and later resamples from this approximation to avoid forgetting. 
Although there are various ways to use the input distribution approximation  (Section~\ref{sec:rel-work}), this study focuses and refers to \emph{rehearsal} in its most direct form, i.e. by sampling the input distribution in a limited rehearsal memory from which samples are revisited in later training batches.

Despite the wide use of rehearsal, due to its simplicity and effectiveness, fundamental analysis of why it works and what its limitations are, is lacking in literature. 
Furthermore, we believe that insights into its internal workings might deepen our understanding of the catastrophic forgetting phenomenon in general.
In this work, we make an initial attempt from the perspective of loss landscapes.

\boldspacepar{Fundamental open questions.}
Motivated by recent advances in continual learning literature, we define two fundamental open questions.
Early work in rehearsal raised concerns about overfitting to the rehearsal memory, as a consequence of repeated optimization on this limited set of data \cite{lopez2017gradient}.
Following work~\cite{chaudhry2019continual} confirms
this as the replay memory becomes perfectly memorized by the model, but also finds rehearsal to remain effective in terms of generalization.
This leads to two open questions.
First, "Why does rehearsal work even though overfitting on the rehearsal memory occurs?".
Second, "How does overfitting on the rehearsal memory influence generalization?".

\boldspacepar{Motivation.}
To formalize these inquiries for this study, we formulate two main hypotheses motivated by advances in prior work.
Firstly, recent work empirically finds continual learning minima of individual tasks to be 
 linearly connected through a low-loss path with the multitask solution, when starting from the same initialization \cite{mirzadeh2021linear}.
 Multitask learning simultaneously learns multiple tasks, which rehearsal ultimately aims to approximate in the continual learning regime. Therefore, we hypothesize the rehearsal solution resides in the same low-loss region as the original task and the multitask solutions.
Secondly, large models are able to completely memorize small sets of data, such as rehearsal memories, without any generalization capabilities \cite{zhang2016understanding}. Therefore we hypothesize that overfitting does harm generalization after all.

\boldspacepar{Contributions.}
Our two hypotheses are formalized as:
\begin{enumerate}
    \item \label{hyp:same-min}  \textbf{Hypothesis 1:}  Rehearsal is effective as its solution tends to reside in \emph{the same} low-loss region as the task minimum from which learning is initiated.
    \item \label{hyp:overfit} \textbf{Hypothesis 2:} Rehearsal is suboptimal as it tends to overfit on the rehearsal memory, consequently harming generalization.
\end{enumerate}
Section~\ref{sec:analysis} provides extensive empirical evidence on MNIST, CIFAR, and Mini-Imagenet data sequences to test both hypotheses.
Our findings can be summarized as follows:
\begin{enumerate}
    \item  The results in Section~\ref{sec:hyp1} unanimously support Hypothesis 1. Even after learning longer sequences of up to 5 tasks with rehearsal, the new minimum is found in the same low-loss region as for the first task. This suggests the existence of overlapping low-loss regions for the task's input distribution and its approximation by the rehearsal memory. 
    \item As suggested in \cite{lopez2017gradient,chaudhry2019continual}, Section~\ref{sec:hyp2} confirms overfitting on the rehearsal memory. However, this overfitting by itself is insufficient to explain harming generalization, as following Hypothesis~\ref{hyp:same-min} there exists an overlapping low-loss region with the one of the task's input distribution. 
    Our empirical evidence finds clues that internal rehearsal dynamics draw the learning trajectory near a high-loss ridge of the rehearsal memory. Additionally, near the high-loss ridge the approximation by the rehearsal memory for the task's input distribution loss deteriorates.
    Therefore, we can confirm the suboptimality of rehearsal in Hypothesis 2, but generalization is harmed by the combination of overfitting, internal rehearsal dynamics, and the low-quality approximation of the rehearsal memory near its high-loss ridges.
\end{enumerate}
Furthermore, Section~\ref{sec:step-back} provides additional evidence with a simple heuristic to withdraw from the rehearsal memory high-loss ridge, showing promising results compared to standard rehearsal.
Additionally, Section~\ref{sec:conceptual-analysis} conceptually analyses the internal rehearsal dynamics w.r.t. our results.

We believe these findings bring new insights into ultimately understanding catastrophic forgetting and proposed methods in literature.
In Section~\ref{sec:revisiting-SOTA} we discuss recent salient works in the light of our findings, namely GEM~\cite{lopez2017gradient}, MIR~\cite{aljundi2019online}, GDumb\cite{prabhu2020gdumb} and linear mode connectivity~\cite{mirzadeh2021linear}.

\section{Related Work}
\label{sec:rel-work}

\boldspacepar{Continual learning.}
Learning from non-stationary data streams has been well studied in literature. Following \cite{9349197}, we divide the main approaches to avoid catastrophic forgetting into three main families: regularization, parameter isolation, and rehearsal based methods. 

Regularization based methods %
put regularization constraints on the parameter space when learning future tasks to preserve acquired knowledge. By using knowledge distillation~\cite{hinton2015distilling}, Li et al.~\cite{li2016learning} insist on staying close to the model distribution from before learning on the new data began.
Other regularization methods \cite{kirkpatrick2017overcoming,aljundi2018memory,zenke2017continual} capture parameter importance based on second-order approximations in task minima. When learning a new task, the squared $l2$-distance to the previous task minimum is minimized, weighed by importance.

Parameter isolation methods allocate subsets of the model parameters to specific tasks.
The allocation ranges from pruning based heuristics \cite{Mallya2017} and learnable masks \cite{Mallya2018,Serra2018} to instantiating new subnetworks per task~\cite{rusu2016progressive}.

\boldspacepar{Rehearsal in continual learning} compresses the observed data in the non-iid data stream and is also called replay or experience replay. One way is by storing a subset of representative samples, called exemplars, in a constrained rehearsal memory~\cite{lopez2017gradient,Rebuffi2017}.
Another way is by learning the input distribution with a generative model~\cite{DGR,shin2017continual,seff2017continual}, referred to as pseudo-rehearsal. 
This approximation of the input distribution is then used when observing new task data to compensate for the non-iid nature of the data stream.
Although promising results for generative models have been reported in confined setups\cite{van2020brain}, learning both the encoder and decoder in a continual fashion can be cumbersome. In comparison, sampling is a more computationally efficient process to approximate the input distribution, avoiding optimization of an additional decoder.

For the rehearsal memory, different approaches to exploit the exemplars have been explored. 
A first straightforward approach is Experience Replay (ER)~\cite{chaudhry2019continual} adding a batch of exemplars to each new batch of data. The union of these is then optimized for the same objective.
In contrast, the exemplars can also be used for knowledge distillation~\cite{Rebuffi2017} or both distillation and the original objective \cite{castro2018end}.
In the latter two works, the exemplars are stored based on optimally representing the feature mean, which is an exhaustive operation and requires recalculating the class means at task boundaries for nearest-mean classification ~\cite{Rebuffi2017}. Therefore, De Lange et al.~\cite{de2020continual} propose an online alternative with continually representative class means in the evolving embedding space.

A second approach to exploit the exemplars, is to consider them to impose constraints in the gradient space~\cite{lopez2017gradient}. The gradient of the new task data is confined to point in the same direction as the task-specific gradients, which are calculated based on the exemplars.
Whenever the gradient fails to satisfy the constraints, it is projected to the least-squares solution on the constraining polyhedral convex cone. 
The gradient space has also been proposed to select samples for storage in the ER rehearsal memory~\cite{aljundi2019gradient}.

As it is infeasible to repeatedly rehearse the entire rehearsal memory, a retrieval strategy is required to select the exemplars for ER.
Random retrieval has been widely adopted~\cite{chaudhry2019continual,chaudhry2018efficient,de2020continual,chrysakis2020online} while an alternative evaluates the potentially highest increase in the loss \cite{aljundi2019online}.

Recent work has shown that storing balanced greedy subsets of the data stream attains major increases in performance when learning offline with the rehearsal memory \cite{prabhu2020gdumb}. This indicates the difficulty of learning in a continual fashion compared to offline learning and suggests the undiscovered potential of fully exploiting the exemplars to learn continually.
We discuss the retrieval and storage strategies for the rehearsal memory in more detail in Section~\ref{sec:CL:retrieval-storage}.

\section{Rehearsal in Continual Learning}
Before considering our analysis, we outline the continual learning setup. Next, Section~\ref{sec:CL:stoch-optim} provides an interpretation of the loss functions in the parameter space in continual learning and compares this to the iid setting. Finally, Section~\ref{sec:CL:retrieval-storage} formally introduces rehearsal and the approximations made during learning.

Typically, machine learning models rely on sampling from a stationary data distribution for both the learner and evaluator. In continual learning, the learner distribution can change over time~\cite{de2020continual}. Specifically in this work, we consider changes in the learner distribution $\mathcal{D}$ at discrete time steps. 
The evaluator distribution $D_{eval}$ is constant and equal to the unison of all learner distributions, with samples drawn mutually exclusive w.r.t. $\mathcal{D}$. This setup is also referred to as the class-incremental setup, 
defined as a sequence of $N$ task sets $D_t$. Each set $D_t$ consists of a set of samples $(\textbf{x}_i, y_i)$, with input data $\textbf{x}_i$ and class label $y_i$. $T1$ and $T2$ refer to the first two tasks, with future tasks referenced similarly.

\subsection{Stochastic optimization in Continual Learning}
\label{sec:CL:stoch-optim}
Standard in machine learning is to optimize parameters $w$ of a predicting function $f_{w}$ w.r.t.~the loss function $\mathcal{L}$. Ideally, the risk of the entire data generating distribution is minimized for $f_{w}$. However, this distribution is unknown and we only have the training set available to learn from, which we use to minimize the empirical risk $R$:
\begin{equation}
\label{eq:risk}
    R\left(\mathcal{D},w \right) = \frac{1}{|\mathcal{D}|} \sum_{\left({\bf x}_i, y_i \right) \in \mathcal{D}} \mathcal{L} \left(f_w({\bf x}_i), y_i \right) 
\end{equation}
The most straightforward way to optimize is by taking steps in the direction of steepest descent, indicated by the negative gradient $g$ of the empirical risk.
However, to maintain scalability for large datasets the stochastic gradient  $\Tilde{g}$ is calculated on a mini batch $B$ from the full dataset. This approximation provably converges, as long as the expected stochastic gradient equals the full set's gradient \cite{bottou2010large}. 
The expected value of $\Tilde{g}$ equals
\begin{equation}
\mathbb{E}\left[ \ \Tilde{g} \ \right] = \sum_{\left({\bf x}_i, y_i \right) \in \mathcal{D}} p_i \ \nabla \mathcal{L} \left(f_w({\bf x}_i), y_i \right) 
\label{eq:expected_gradient}
\end{equation}
with $p_i=S\left(d_i \right)$ the sampling probability for sample ${d_i  \in \mathcal{D}}$. 
For a uniform sampling distribution, the expectation of $\Tilde{g}$ becomes equal to $g$.

However, in continual learning $p_i$ is non-uniform. In our setting, it is only non-zero and uniformly distributed over samples belonging to the current task.
Reformulating Eq.\ \ref{eq:expected_gradient} for task datasets $\mathcal{D}_{1..N}$ and current task $\mathcal{D}_c$ reduces to
\begin{equation}
\begin{split}
\mathbb{E}\left[ \ \Tilde{g} \ \right] &= \sum_{\mathcal{D}_t} \sum_{\left({\bf x}_i, y_i \right) \in \mathcal{D}_t} p_i \ \nabla \mathcal{L} \left(f_w({\bf x}_i), y_i \right) \\
&= \sum_{\left({\bf x}_i, y_i \right) \in \mathcal{D}_c} p_i \ \nabla \mathcal{L} \left(f_w({\bf x}_i), y_i \right) \ 
\end{split}
\label{eq:expected_gradient_tasks}
\end{equation}
As a consequence,
when training on task distribution $\mathcal{D}_c$, optimization disregards the other task distributions and calculates the gradients solely from the perspective of the current task loss function. This is the key issue resulting in catastrophic forgetting of previous tasks in neural networks.

\subsection{Learning with rehearsal}
\label{sec:CL:retrieval-storage}

Continual learning systems typically have fixed memory budgets and preferably do not grow with the number of tasks.
Consequently, this requires making a trade-off with rehearsal methods approximating the input distribution instead of storing all observed data.
The operational memory $\mathcal{M}$ defines the additional memory requirements during learning of the continual learner\cite{de2020continual}.
Rehearsal implements $\mathcal{M}$ as a fixed rehearsal memory, with the memory for task distribution $\mathcal{D}_t$ indicated by $\mathcal{M}_t$. The expected gradient in Eq. \ref{eq:expected_gradient_tasks} then sums over the union of both the current task data $\mathcal{D}_c$ and the stored samples in $\mathcal{M}$.
This union could also be constructed with a generative model for pseudo-rehearsal as discussed in Section~\ref{sec:rel-work}, but the scope of this study is limited to rehearsal by sampling.

\begin{figure*}[!ht]
\centering
\begin{subfigure}{.5\linewidth}
\includegraphics[clip,trim={0cm 0.5cm 0cm 0cm},width=1\linewidth]{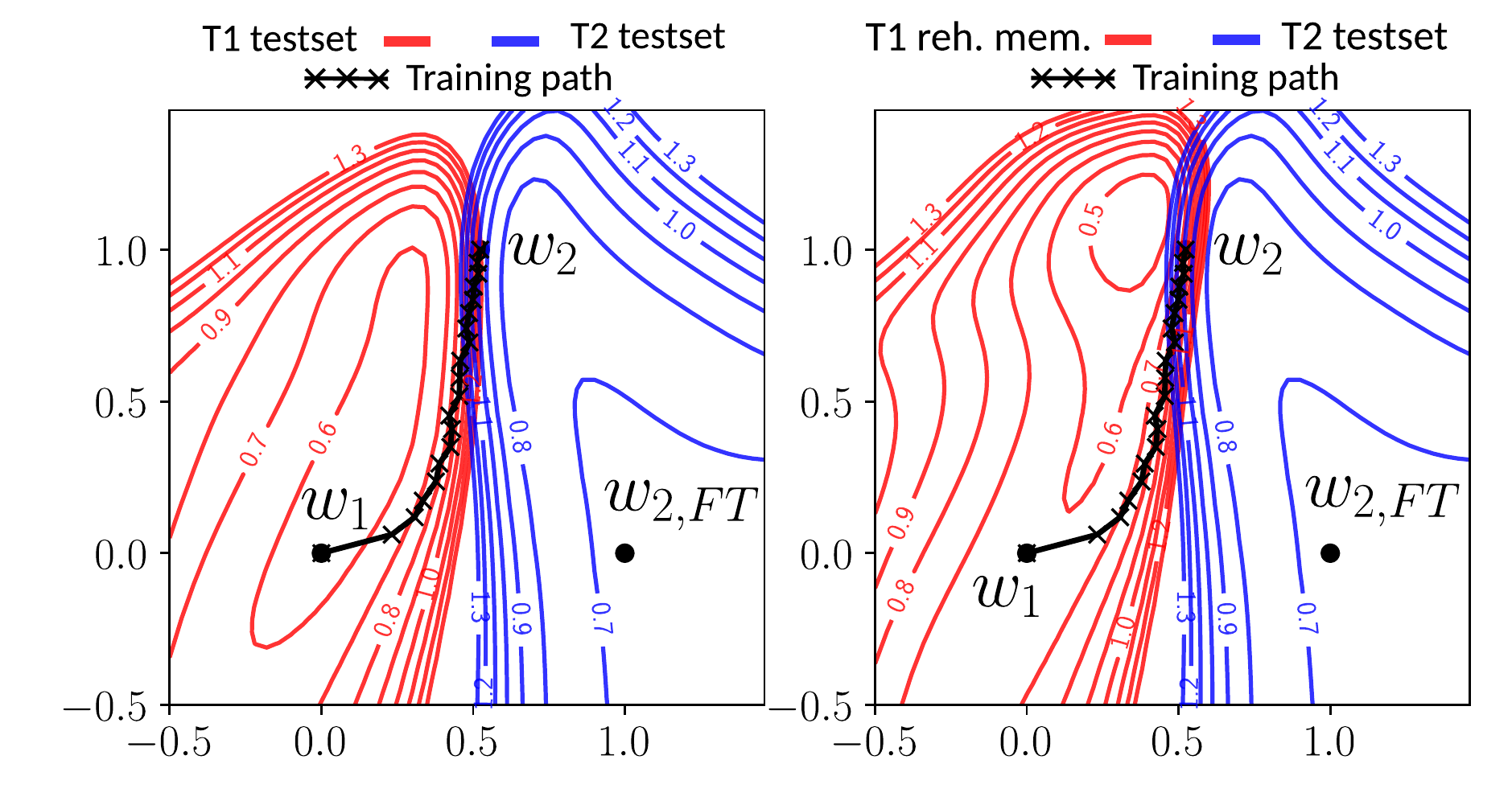}
\caption{Loss function on CIFAR10}
\end{subfigure}%
\begin{subfigure}{.5\linewidth}
\includegraphics[clip,trim={0cm 0.5cm 0cm 0cm},width=1\linewidth]{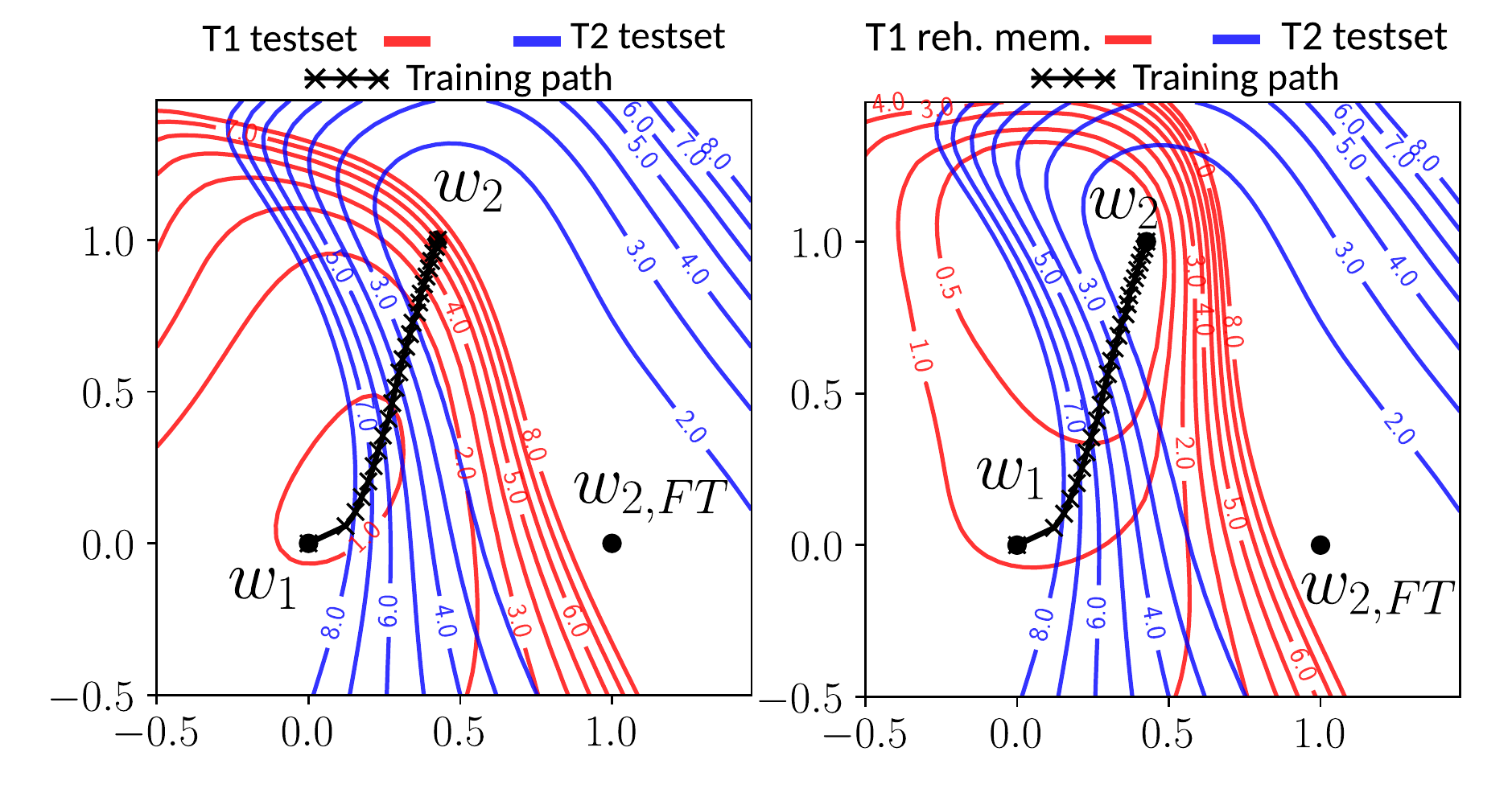}
\caption{Loss function on Mini-Imagenet}
\end{subfigure}
   \caption{Projection of learning trajectories in parameter space on the plane of $w_1$, $w_2$ and $w_{2,FT}$.
   For the same $T2$ loss (blue), the loss for $T1$ (red) is calculated in two different ways.
   \emph{(a) and (b) left}: $T1$ loss for the vast test set. \emph{(a) and (b) right}: $T1$ loss for the limited rehearsal memory.
   Even in the 2D planes, overfitting to the rehearsal memory loss is clear. 
  See Appendix for details.
   }
\label{fig:contour_plots}
\end{figure*}

\begin{figure*}[!ht]
    \centering
    \includegraphics[clip,trim={0cm 0cm 0cm 0cm},width=\linewidth]{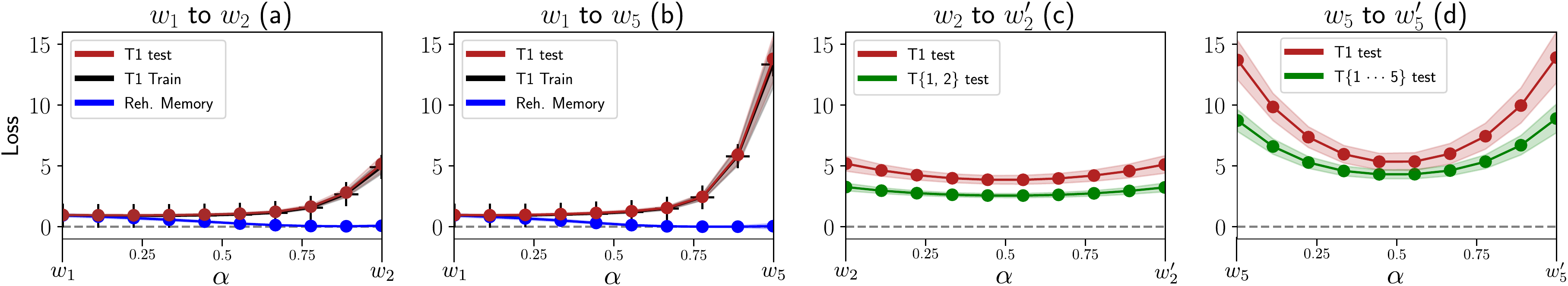}
    \caption{Avg.\ loss and standard deviation on linear paths between the models declared in Figure \ref{fig:setup}, trained on Mini-Imagenet and sampled 100 times for different model initializations and memory populations.
    A path from $w_i$ to $w_j$ is calculated as $(1-\alpha)w_i + \alpha w_j$. 
    \emph{(a) and (b)}: %
    Loss on the linear path between the model after learning $T1$ ($w_1$) and the model after learning with rehearsal on $T2$ and $T5$ respectively. 
    \emph{(c) and (d)}: %
    Loss of the path between two models learned with different memory populations, after $T2$ and $T5$ respectively. Red is the loss on the T1 testset, green the average loss of all tasks up to T2 and T5 respectively.
     Results contained no outliers with a higher loss on the path compared to the loss of the models.
    }
    \label{fig:paths:MIMG}
\end{figure*}

Rehearsal is confined by two main constraints.
Firstly, the limited rehearsal memory size $|\mathcal{M}|$ to approximate the input distribution, and secondly, the limited mini batch size $|B|$ in SGD to construct near iid mini batches that approximate Eq.\ \ref{eq:expected_gradient}.
Selecting samples from the input distribution to store in $\mathcal{M}$ is handled by the \emph{storage policy}, whereas samples for mini batch $B$ are selected based on sampling distribution $S$, defined by the \emph{retrieval policy}. Algorithm~\ref{algor:rehearsal} summarizes rehearsal for a single mini batch $B$.

In the following, we assume the full memory $\mathcal{M}$ to be equally divided over all seen tasks with a random task population. The retrieval policy's sampling distribution is uniform over $\mathcal{M}$.

\begin{algorithm}[H]
\begin{algorithmic}[1] %
\Function{RehearsalBatch}{$B, \mathcal{M}$}
\State $\Tilde{B} \leftarrow$\textsc{RetrievalPolicy}$(\mathcal{M})$ \Comment{{\footnotesize Retrieve exemplars}}
\State $w \leftarrow SGD\left( B \cup \Tilde{B}, w  \right)$ \Comment{{\footnotesize Optimize objective for union}}
\State \textsc{StoragePolicy}$(\mathcal{M}, B)$ \Comment{{\footnotesize Update rehearsal memory}}
\EndFunction
\end{algorithmic}
\caption{\label{algor:rehearsal}Continual learning with Rehearsal. }
\end{algorithm}

\section{Revealing Rehearsal: Analysis}
\label{sec:analysis}

Although rehearsal has been widely adopted in continual learning literature, the proposed approaches are often based on heuristics. In contrast, in this section we attempt to gain fundamental insights from the perspective of the loss landscape in parameter space which is also key to progress our understanding of catastrophic forgetting in general.

In the following, we define $w_1$ as the minimum obtained by the model after learning until convergence for the first task ($T1$) with task distribution $\mathcal{D}_1$. Subsequent models $w_i$ are learned via rehearsal as defined in Section~\ref{sec:CL:retrieval-storage}, e.g. $w_2$ is the solution for $T2$ initialized from $w_1$ and using samples from $T1$ stored in $\mathcal{M}$ during learning. 
We define a low-loss region as a connected part of the parameter space where the loss value stays below a small value. Any set of parameters in such a region will perform nearly equally well. Such regions are an important part of the parameters space because they are linked to better generalization capabilities of a model \cite{izmailov2018averaging, keskar2016large}.

Our two hypotheses are formulated in Section~\ref{sec:intro}.
We scrutinize the hypotheses in Section~\ref{sec:hyp1} and Section~\ref{sec:hyp2} respectively with strong empirical evidence on three continual learning benchmarks and discuss the consequences with relation to hypotheses in previous work.
Using these insights, we exemplify a simple baseline in Section~\ref{sec:step-back} further confirming our hypotheses.
Section~\ref{sec:conceptual-analysis} conceptually analyses the observed rehearsal dynamics. 
Finally, Section~\ref{sec:comparison-prior-evidence} compares our observations with prior work.

\boldspacepar{Datasets.}
We use three datasets for the rehearsal analysis: Split-MNIST, Split-CIFAR10, and Split-miniImagenet. Split-MNIST divides the original MNIST \cite{mnist} database into 5 tasks, with each 2 classes. Split-CIFAR10 uses the CIFAR10 dataset \cite{cifar}, and is similarly split into five two-label tasks. Split-miniImagenet is constructed by dividing miniImagenet~\cite{miniimagenet}, a 100-class subset of Imagenet, into 20 tasks with 5 labels each. For the analyses, only the first five tasks (i.e. 25 classes) are used in both training and evaluation. Unless mentioned otherwise, the MNIST and CIFAR10 sets are trained online, which implies that except for the data in memory, all data is seen only once. Split-miniImagenet uses 10 epochs per task since it is a much harder dataset.
For brevity, we refer to the setups as MNIST, CIFAR10, and Mini-Imagenet.

\boldspacepar{Architectures and optimization.} MNIST is trained on a fully-connected network, with two hidden layers of 400 nodes each. Both CIFAR10 and Mini-Imagenet are trained on a reduced Resnet18~\cite{he2016deep}, introduced by Lopez-Paz et al.\ \cite{lopez2017gradient}. All networks are trained with a shared final layer, referred to as shared head. This is opposed to the easier task-incremental setting which uses a different head per task \cite{9349197}. The memory sizes for the three datasets are respectively 50, 100, and 100 samples per task for MNIST, CIFAR10, and Mini-Imagenet. All models are optimized using vanilla SGD with a cross-entropy loss. 
Each mini batch during training consists of 10 new and 10 memory samples, except for the first task which only has 10 new samples. 
Our code uses the Avalanche framework~\cite{lomonaco2021avalanche} to enhance reproducibility.
We summarize all details in Appendix.

\subsection{Hypothesis 1: Empirical evidence}
\label{sec:hyp1}
To test Hypothesis~\ref{hyp:same-min}, we need to quantify whether the model before and after learning a new task remains in the same low-loss region.
For clarity, we start with an example of two subsequent tasks: after learning the first task ($w_1$), we can either learn $T2$ using rehearsal ending up in $w_2$, or finetune for $T2$ without rehearsal to end up in a $T2$ minimum ($w_{2,FT}$) while typically catastrophically forgetting $T1$. 
Figure~\ref{fig:contour_plots} shows a two-dimensional projection in the parameter space on the plane defined by these three models, for CIFAR10 and Mini-Imagenet (see Appendix for MNIST results). 
 The projection of the learning trajectory in parameter space
in Figure~\ref{fig:contour_plots}
illustrates the large steps initially moving towards the $T2$ minimum $w_{2,FT}$, to then bend with smaller steps along the high-loss contour lines of $T1$'s rehearsal memory loss landscape.
These findings indicate $w_1$ and $w_2$ remaining in the same low-loss region, while $w_2$ is drawn near the high-loss ridge of the rehearsal memory. 

To further support these findings, we analyze the loss in Figure~\ref{fig:paths:MIMG} both after the second task ($w_2$) and after a sequence of 5 tasks ($w_5$) of Mini-Imagenet.
Focusing on the loss values of $T1$ on the linear path from $w_1$ to $w_2$ (Figure~\ref{fig:paths:MIMG}a) and from $w_1$ to $w_5$ (Figure~\ref{fig:paths:MIMG}b), we observe monotonically increasing loss values for $T1$, hence indicating the compared models are in or on the edge of the same loss basin.
The results are averaged over 100 runs with both different initialization and rehearsal memory population.
We refer to Appendix for CIFAR10 and MNIST results, having the same trends as for Mini-Imagenet.
Conform to literature, rehearsal is effective in alleviating catastrophic forgetting, as it significantly improves results compared to plain finetuning with $60\%$, $9\%$ and $7\%$ gain in average accuracy over all tasks for the increasingly difficult MNIST, CIFAR10, and Mini-Imagenet benchmarks.

Another question that arises within this analysis is: \emph{"Do these findings still hold for different rehearsal memory populations
and how are their solutions connected?"} 
To answer this, we extend the previous analysis by considering two alternative memory populations resulting in $w_2'$ and $w_5'$. 
Figure~\ref{fig:paths:MIMG} plots the loss values for the linear interpolation between the models $w_2$ and $w_2'$, and $w_5$ and $w_5'$. As there is no increase in the loss value, we can conclude that for different random memory populations the resulting models all reside in the same low-loss basin as $w_1$. 
Figure~\ref{fig:setup} summarizes the setup for the $T2$ paths (a) and (c) in Figure \ref{fig:paths:MIMG}, with paths (b) and (d) defined analogously after 5 tasks.

\begin{figure}[!ht]
\centering
\includegraphics[clip,trim={0cm 0.1cm 0cm 0cm},width=1\linewidth]{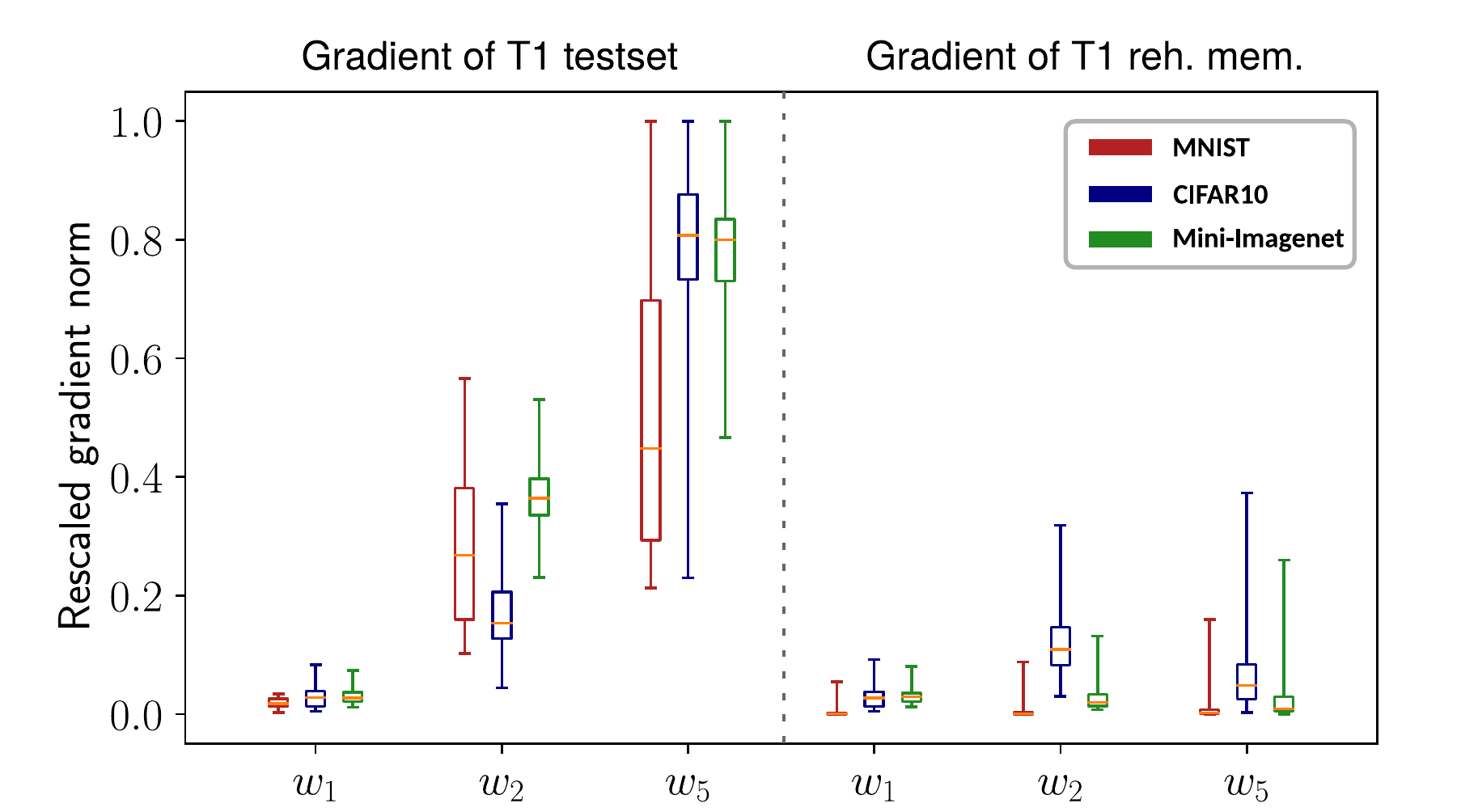}
   \caption{\label{fig:gradientnorms}
   Comparison of rescaled gradient $l2$-norms of the $T1$ testset (left) and rehearsal memory (right) 
   after learning $T1$ ($w_1$), $T2$ ($w_2$) and $T5$ ($w_5$), averaged over $100$ seeds.
    Boxplot whiskers indicate \emph{max} and \emph{min} values.
   Results are rescaled by the maximal norm value per dataset.
   }
\label{fig:long}
\end{figure}

\subsection{Hypothesis 2: Empirical evidence}
\label{sec:hyp2}

Testing Hypothesis~\ref{hyp:overfit} requires an experiment showing overfitting on the rehearsal memory with deteriorating generalization as overfitting arises.
We expand the initial two-task experiment for Hypothesis~\ref{hyp:same-min} by analyzing the optimization trajectory in the loss landscape of both the rehearsal memory and the test dataset.
Starting with the loss landscape for the rehearsal memory of $T1$, Figure~\ref{fig:contour_plots}a (right) and Figure~\ref{fig:contour_plots}b (right) depict the rehearsal minimum $w_2$ to be in a low-loss region next to a high-loss ridge.
However, in perspective of the loss landscape computed on the full test set in Figure~\ref{fig:contour_plots}a (left) and Figure~\ref{fig:contour_plots}b (left), $w_2$ resides \emph{on} the high-loss ridge. This indicates that overfitting on the loss landscape for the rehearsal memory is harmful for generalization, especially as rehearsal draws the $w_2$ solution towards the high-loss ridge. 

We confirm these findings further for longer sequences of tasks and observe for Mini-Imagenet in Figure~\ref{fig:paths:MIMG}a and Figure~\ref{fig:paths:MIMG}b that overfitting occurs for the rehearsal memory as indicated by the zero loss for $w_2$ and $w_5$. Moreover, these results suggest the harmfulness for generalization as $w_2$ and $w_5$ locate on the high-loss ridge for the $T1$ test dataset.

Furthermore, we find the rehearsal memory's loss to provide a poor approximation of $T1$'s high-loss ridge of the full training data.
In Figure~\ref{fig:paths:MIMG}a and b the loss on the linear paths from $w_1$ to $w_2$ and $w_5$ is similar near $w_1$ but increases significantly near $w_2$ and $w_5$ for the $T1$ training data, while the rehearsal memory overfits.
This is problematic, because rehearsal can only observe the rehearsal memory's view of the loss landscape. 
The rehearsal dynamics draw the solution near a high-loss ridge for the rehearsal memory and this is where the high-loss ridge is being poorly approximated.
Therefore, instead of ending up \emph{near} the high-loss ridge in perspective of the rehearsal memory, the solution in reality resides \emph{on} the high-loss ridge for the training data, which consequently also harms generalization.

Additionally, we analyze the gradient norms for the $w_1$, $w_2$ and $w_5$ minima for all three benchmarks in Figure~\ref{fig:gradientnorms}. %
By quantifying the gradient norms for both the test dataset and the rehearsal memory for $T1$, we can confirm how throughout learning the sequence of tasks, the solution resides in a low-loss region for the rehearsal memory. However, the gradient norms on the test dataset increase from $w_1$ to $w_2$ and significantly increase further as more tasks have been learned ($w_5$). This indicates that as the number of learned tasks increases, more overfitting occurs on the rehearsal memory and generalization deteriorates.

\subsection{High-loss ridge aversion}
\label{sec:step-back}
The previous two sections report strong empirical evidence on three benchmarks to support our two hypotheses.
In this section, we conduct an additional experiment relying on these insights to avoid the overfitting towards the high-loss ridges of the rehearsal memory.

Considering two tasks, from Hypothesis~\ref{hyp:same-min} we can derive that the $w_2$ minimum resulting from rehearsal with $\mathcal{M}_1$ resides in the same low-loss region as $w_1$. 
Therefore, it follows that the low-loss region of the rehearsal memory $\mathcal{M}_1$ at least has an overlap with the low-loss region of $\mathcal{D}_1$.
Additionally, Hypothesis~\ref{hyp:overfit} shows $w_2$ overfits on the rehearsal memory $\mathcal{M}_1$, with our empirical evidence in Section~\ref{sec:hyp2} illustrating the solution being drawn close to a high-loss ridge of $\mathcal{M}_1$.

Therefore, we test a simple heuristic to withdraw from the high-loss ridges, by isolating training on the rehearsal memory for $n$ updates after converging to the new task minimum $w_2$. As this gives incentive for an inward movement in the $\mathcal{M}_1$ low-loss region, we also expect inward movement for the overlapping $T1$ low-loss region. This would reduce overfitting on the rehearsal memory and hence improve generalization.
Table~\ref{tab:step-back} shows the results for different memory sizes $|\mathcal{M}|$ for sequences of 5 tasks.
For MNIST with the smallest rehearsal memory of 100 exemplars, we observe a significant improvement in generalization of $9.4\%$, reducing to $1.6\%$ margin for 10 times more exemplars.
These observations are aligned with our expectations, as smaller rehearsal memories suffer more from overfitting and therefore gain more by withdrawing from the high-loss ridge.
The margin of increased generalization declines for the more difficult CIFAR10 and Mini-Imagenet sequences. Nonetheless, both sequences attain a respective margin of $0.7\%$ and $4.1\%$ for a small memory of $100$ samples. 
To clearly demonstrate the effect of high-loss ridge aversion, all benchmarks train for 10 epochs per task to encourage overfitting on the rehearsal memory. For the two more challenging setups Stable-SGD~\cite{mirzadeh2020understanding} was used to attain wider minima.

\begin{table}[!htbp]
\centering
\resizebox{\linewidth}{!}{%
\begin{tabular}{@{}llll@{}}
\toprule
\textbf{Experiment}  & \multicolumn{3}{c}{\textbf{Rehearsal memory size $ \bf |\mathcal{M}|$}}                                   \\ \midrule
                     & \textit{100}          & \textit{500}          & \textit{1000}         \\
\textbf{MNIST} &                       &                       &                       \\
\textit{ER}          & $72.9 \pm 1.8$        & $87.1 \pm 0.6$        & $90.2 \pm 0.5$ \\
\textit{ER-step} & $82.3 \pm 2.2$ (n=20) & $90.0 \pm 0.4$ (n=10) & $91.8 \pm 0.3$ (n=10)\\
\addlinespace
\textbf{CIFAR10}     &              &              &               \\ 
\textit{ER (stable)}         & $46.7 \pm 0.8$        & $65.1 \pm 1.0$        & $70.7 \pm 1.7$       \\
\textit{ER-step (stable)} & $47.4 \pm 1.9$ (n=10) & $66.2 \pm 1.1$ (n=10) & $70.8 \pm 1.0$ (n=1) \\ \addlinespace
\textbf{Mini-Imagenet}     &              &              &               \\ 
\textit{ER (stable)}          & $25.7 \pm 1.7$        & $39.7 \pm 1.0$        & $46.6 \pm 1.1$        \\
\textit{ER-step (stable)} & $29.8 \pm 2.0$ (n=20) & $42.3 \pm 1.1$ (n=10) & $47.4 \pm 1.4$ (n=10) \\
\bottomrule
\end{tabular}
}
\caption{\label{tab:step-back} Avg. acc. for $D_{eval}$ after learning a sequence of 5 tasks for different rehearsal memory sizes $|\mathcal{M}|$. \emph{ER-step}: After attaining each task minimum, $n$ update steps are performed solely on rehearsal memory, compared to \emph{ER} (ER-step with $n=0$). 
Stable indicates using Stable-SGD~\cite{mirzadeh2020understanding}.
We report standard deviation over 5 initializations.}
\end{table}

\subsection{Conceptual analysis of rehearsal dynamics}
\label{sec:conceptual-analysis}
The results in Section~\ref{sec:hyp1} show that during rehearsal, the learning trajectory is initially drawn to a low-loss region of the new task before deflecting towards a minimum near the rehearsal memory's high-loss ridge.
This can be explained by the relation between the loss values and the gradient magnitude. Following \cite{farajtabar2020orthogonal}, the gradient of the loss $\nabla \mathcal{L}$ can be decomposed using the chain rule into  
the gradient of the model $\nabla f_w$ w.r.t. its parameters $w$ and the derivative of the loss $\mathcal{L}'$ w.r.t. the output of $f_w$:
\begin{equation}
    \nabla \mathcal{L}(f_w(\textbf{x}), y) = \nabla f_w(\textbf{x}) \ \mathcal{L}'(y, f_w(\textbf{x})) \ .
\end{equation}
For the standard cross-entropy loss with softmax $p(\cdot)$, the derivatives w.r.t. the elements of the output ${\bf \hat{y}} = f_w(\textbf{x})$ become
$\mathcal{L}'=\left(p(\hat{y}_c) - 1 \right)$ for $\hat{y}_c$ corresponding to the ground-truth and $\mathcal{L}'= p(\hat{y}_i)$ for the other output elements ${\hat{y}_i,} {\forall i \neq c}$. 
This means that the gradients will be smaller if the output is closer to the ground truth one-hot vector. As shown in Eq.\ \ref{eq:expected_gradient_tasks}, the final gradient is the average of the gradients on the individual tasks. At $w_1$, the exemplars of $T1$ will have close to zero loss, while the new samples of $T2$ typically start at a high loss. 
Therefore, the gradient direction and magnitude are initially dominated by $T2$. However, as observed in Figure~\ref{fig:contour_plots}, the loss for $T1$ increases until both task losses are balanced, and the trajectory continues on loss contours of similar magnitude.

\subsection{Comparison to prior evidence}
\label{sec:comparison-prior-evidence}
In this section, the empirical findings supporting our hypotheses are compared to empirical evidence found in prior work.
Concerns about overfitting on the rehearsal memory have been expressed in prior studies \cite{lopez2017gradient}. 
This was later criticized in up-following work \cite{chaudhry2019continual}, with their hypothesis formulated as
\emph{``although direct training on the examples
in $\mathcal{M}_1$ (in addition to those coming from [$T2$]) does indeed lead to strong memorization of $\mathcal{M}_1$ [...], such training is still overall beneficial in terms of generalization on the original task $T1$ because the joint learning with the examples of the current task $T2$ acts as a strong, albeit implicit and data-dependent, regularizer for $T1$."}
Their empirical evidence is based on a minimal MNIST rotation experiment where two subsequent tasks are considered for different levels of relatedness, i.e. Task 1 ($0^{\circ}$) is compared to Task 2 for different degrees of rotation ($20^{\circ}, 40^{\circ}, 60^{\circ}$).
For all levels of relatedness, they find training $T2$ with a small $T1$ rehearsal memory $\mathcal{M}_1$ remains beneficial for the $T1$ performance.

With the empirical findings in our study, we are able to explain \emph{why} learning with a small rehearsal memory remains beneficial for generalization. However, we also indicate how overfitting in rehearsal can become harmful for generalization, hence providing counter-evidence for their hypothesis. Furthermore, our findings are explanatory of how learning $T2$ in their experiments can exhibit regularizing effects when $T2$ is highly related to $T1$.

Firstly, we address why learning with a small rehearsal can still generalize. In Section~\ref{sec:hyp1} we show that the low-loss regions of $T1$ and its rehearsal memory $\mathcal{M}_1$ intersect after learning $T1$. Our findings suggest that learning $T2$ with rehearsal remains in the same low-loss region of $\mathcal{M}_1$. This indicates that although overfitting can happen, a generalizing solution can be found because of the overlap between both regions.

Secondly, we address how overfitting in rehearsal can become harmful for generalization. The experiments in Section~\ref{sec:hyp2} clearly exhibit overfitting on the rehearsal memory.
This finding by itself doesn't explain how overfitting becomes harmful for generalization, as we just indicated there exists a generalizing solution in the overlapping low-loss region for $T1$ and $\mathcal{M}_1$.
However, Section~\ref{sec:hyp2} finds the rehearsal solution ending up near a high-loss ridge, where the approximation of the loss-landscape of $\mathcal{M}_1$ for $\mathcal{D}_1$ deteriorates.
Therefore, our results suggest it is the dynamic of both overfitting and the rehearsal solution ending up near a high-loss ridge harming generalization.

As a consequence, thirdly, for dissimilar tasks $T2$ has no regularization effects for $T1$, and on the contrary, deteriorates performance by pulling $w_2$ towards the $\mathcal{M}_1$ high-loss ridge.

\section{Revisiting state-of-the-art}
\label{sec:revisiting-SOTA}
In this section, we will interpret the successes and results of the state-of-the-art in continual learning in the light of the insights in Section~\ref{sec:analysis}. Although progress has been made in the field, a universal and fundamental understanding of these findings is yet to emerge. In the following, we make an initial effort in the perspective of loss landscapes. 

\subsection{Rehearsal methods}

\boldspacepar{GEM}~\cite{lopez2017gradient} constrains the gradient of new samples such that there is no loss increase on the samples in the memory. 
By strictly adhering to the gradient constraint, GEM ensures that the model will reside in the same low-loss basin of the first task. However, the constraints also may encourage overfitting on the rehearsal memory as even within the low-loss basin no slight increase on the loss is allowed. 
In contrast, rehearsal allows increases in the memory's loss as long as it comes with an equal or larger loss decrease in the new task (proof in Appendix). This makes wider exploration of the low-loss basin possible compared to a model trained with GEM. This might be an indication for the results in favor of rehearsal in \cite{chaudhry2019continual,aljundi2019online,de2020continual,aljundi2019gradient}.

Additionally, finding modified gradients in GEM is computationally expensive, and although improved in \cite{chaudhry2018efficient}, remains costly compared to simple rehearsal.

\boldspacepar{MIR}~\cite{aljundi2019online} defines a retrieval policy for rehearsal by sampling exemplars from $\mathcal{M}$ with the highest increase in loss, measured by a tentative update of the model. 
As Section~\ref{sec:hyp2} indicates, overfitting occurs on the rehearsal memory. 
The MIR retrieval policy increases the sampling probability based on the per-sample loss increase. The loss landscape in Figure~\ref{fig:contour_plots} visualizes the high-loss ridge for the \emph{average} over all rehearsal memory samples of the first task. 
MIR is more likely to select individual samples with the closest high-loss ridge when drawn to the $T2$ low-loss region. This possibly keeps the $w_2$ solution further from the average rehearsal memory high-loss ridge, although with the downside of majorly reselecting the same subset of $\mathcal{M}$, increasing the risk of further overfitting.

\boldspacepar{GDumb}~\cite{prabhu2020gdumb} is a controversial work questioning the progress made in continual learning by proposing a new baseline. Their Greedy Sampler and Dumb Learner (GDumb) greedily stores samples balanced over the observed classes and at inference learns a new model from scratch with the rehearsal memory.
This simple baseline outperforms a vast range of continual learning methods.
For the rehearsal methods, our findings show that the rehearsal minimum resides in the same low-loss region as $T1$. As more tasks are learned, the overlap of the low-loss regions can only decline for the union of all tasks. Therefore, it becomes gradually harder to find a solution for all tasks in the sequence.
In contrast, GDumb overcomes this limitation by learning a model from scratch, enabling to find a joint minimum for all rehearsal memories. 

Although this is an effective approach to combat the 
narrowing parameter subspace with low loss for all tasks,
a concern regarding the GDumb baseline is to what extent it can be regarded as a continual learner.
The learning process is repeated for each task from scratch and doesn't build on the knowledge base acquired in the model learned on previous tasks. In other words, there is no transfer learning involved between subsequent tasks, reducing it to a sampling problem for the rehearsal memory as \emph{learning} doesn't happen continually.

\subsection{Multitask mode connectivity}

Linear mode connectivity~\cite{mirzadeh2021linear} aims to find the relation between the multitask solution $w_{1,2}$, by simultaneously learning $T1$ and $T2$, with a sequentially learned solution $w_{2,FT}$, conditioned by starting from the same initialization $w_1$. 
Their empirical evidence suggests the multitask solution $w_{1,2}$ to be connected with both $w_1$ and $w_{2,FT}$ through linear low-loss paths.
This has a significant consequence for rehearsal in continual learning. 
First of all, once a minimum $w_1$ is found for the first task, we can assume that for future tasks, there exists an overlapping parameter subspace of low loss within the low-loss region of $w_1$ where the multitask solution resides.
In rehearsal, instead of jointly training on all observed tasks, a subset of observed samples is stored in the rehearsal memory. Therefore, it can be seen as an approximation of the multitask objective. 
As our empirical findings confirm in Section~\ref{sec:hyp1}, the rehearsal solution $w_2$ indeed resides within the same low-loss region as $w_1$.

\subsection{Loss-based Parameter Regularization}
Related work in Section~\ref{sec:rel-work} introduced another important family of \emph{regularization-based methods} in continual learning.
In this section we compare rehearsal to pioneering work in this family, Elastic Weight Consolidation~\cite{kirkpatrick2017overcoming}, inspiring many other works \cite{zenke2017continual,aljundi2018memory,liu2018rotate,lange2020unsupervised}.
Their approach makes a second-order approximation in the task minimum to constrain learning of further tasks through regularization, i.e. directions of high loss curvature are discouraged.
However, this is a single-point approximation with quadratic assumptions on the shape of the loss and assumes zero covariance for computational feasibility.
This results in a fixed ellipsoid approximation of the loss contours in the minimum, which is contradicted by our linear interpolation plots in Figure \ref{fig:contour_plots} and other work on loss landscapes~\cite{draxler2018essentially, mirzadeh2021linear}. 
Additionally, any deviation in parameter space from the point of estimation, i.e. the task minimum, possibly results in deteriorating quality of the quadratic loss approximation.
In contrast, rehearsal approximates the entire loss landscape based on sampling the input distribution. 
Therefore, the quality of the approximation is limited by the rehearsal memory but is independent of a point of estimation and no explicit assumptions are made on the loss shape.

\section{Conclusion}
This work investigated the open fundamental questions of \emph{why rehearsal works even though overfitting on the rehearsal memory occurs}, and 
\emph{how this overfitting influences generalization}.
To answer these questions, we formalized two hypotheses in the introduction, which were confirmed by comprehensive empirical evidence on three common benchmarks. Our observations suggest that rehearsal prevents a sequential model from leaving the first found low-loss region, and that this model is susceptible to overfitting towards the edge of the rehearsal memory's low-loss region, harming generalization. 
These findings enhance our understanding in both rehearsal and continual learning dynamics.
We hope to encourage further research in this direction with a focus on other continual learning methods.

\clearpage
\newpage


\begin{thebibliography}{10}\itemsep=-1pt

\bibitem{aljundi2018memory}
Rahaf Aljundi, Francesca Babiloni, Mohamed Elhoseiny, Marcus Rohrbach, and
  Tinne Tuytelaars.
\newblock Memory aware synapses: Learning what (not) to forget.
\newblock In {\em ECCV}, pages 139--154, 2018.

\bibitem{aljundi2019online}
Rahaf Aljundi, Lucas Caccia, Eugene Belilovsky, Massimo Caccia, Min Lin,
  Laurent Charlin, and Tinne Tuytelaars.
\newblock Online continual learning with maximally interfered retrieval.
\newblock {\em Proceedings NeurIPS 2019}, 32, 2019.

\bibitem{aljundi2019gradient}
Rahaf Aljundi, Min Lin, Baptiste Goujaud, and Yoshua Bengio.
\newblock Gradient based sample selection for online continual learning.
\newblock {\em NeurIPS}, pages 11816--11825, 2019.

\bibitem{bottou2010large}
L{\'e}on Bottou.
\newblock Large-scale machine learning with stochastic gradient descent.
\newblock In {\em Proceedings of COMPSTAT'2010}, pages 177--186. Springer,
  2010.

\bibitem{castro2018end}
Francisco~M Castro, Manuel~J Mar{\'\i}n-Jim{\'e}nez, Nicol{\'a}s Guil, Cordelia
  Schmid, and Karteek Alahari.
\newblock End-to-end incremental learning.
\newblock In {\em Proceedings of the European conference on computer vision
  (ECCV)}, pages 233--248, 2018.

\bibitem{chaudhry2018efficient}
Arslan Chaudhry, Marc’Aurelio Ranzato, Marcus Rohrbach, and Mohamed
  Elhoseiny.
\newblock Efficient lifelong learning with a-gem.
\newblock In {\em International Conference on Learning Representations}, 2018.

\bibitem{chaudhry2019continual}
Arslan Chaudhry, Marcus Rohrbach, Mohamed Elhoseiny, Thalaiyasingam Ajanthan,
  Puneet~K Dokania, Philip~HS Torr, and M Ranzato.
\newblock Continual learning with tiny episodic memories.
\newblock 2019.

\bibitem{chrysakis2020online}
Aristotelis Chrysakis and Marie-Francine Moens.
\newblock Online continual learning from imbalanced data.
\newblock In {\em International Conference on Machine Learning}, pages
  1952--1961. PMLR, 2020.

\bibitem{de2020continual}
Matthias De~Lange and Tinne Tuytelaars.
\newblock Continual prototype evolution: Learning online from non-stationary
  data streams.
\newblock {\em arXiv preprint arXiv:2009.00919}, 2020.

\bibitem{9349197}
M. {Delange}, R. {Aljundi}, M. {Masana}, S. {Parisot}, X. {Jia}, A.
  {Leonardis}, G. {Slabaugh}, and T. {Tuytelaars}.
\newblock A continual learning survey: Defying forgetting in classification
  tasks.
\newblock {\em IEEE Transactions on Pattern Analysis and Machine Intelligence},
  pages 1--1, 2021.

\bibitem{draxler2018essentially}
Felix Draxler, Kambis Veschgini, Manfred Salmhofer, and Fred Hamprecht.
\newblock Essentially no barriers in neural network energy landscape.
\newblock In {\em International conference on machine learning}, pages
  1309--1318. PMLR, 2018.

\bibitem{farajtabar2020orthogonal}
Mehrdad Farajtabar, Navid Azizan, Alex Mott, and Ang Li.
\newblock Orthogonal gradient descent for continual learning.
\newblock In {\em International Conference on Artificial Intelligence and
  Statistics}, pages 3762--3773. PMLR, 2020.

\bibitem{french1999catastrophic}
Robert~M French.
\newblock Catastrophic forgetting in connectionist networks.
\newblock {\em Trends in cognitive sciences}, 3(4):128--135, 1999.

\bibitem{he2016deep}
Kaiming He, Xiangyu Zhang, Shaoqing Ren, and Jian Sun.
\newblock Deep residual learning for image recognition.
\newblock In {\em Proceedings of the IEEE conference on computer vision and
  pattern recognition}, pages 770--778, 2016.

\bibitem{hinton2015distilling}
Geoffrey Hinton, Oriol Vinyals, and Jeff Dean.
\newblock Distilling the knowledge in a neural network.
\newblock {\em arXiv preprint arXiv:1503.02531}, 2015.

\bibitem{izmailov2018averaging}
Pavel Izmailov, Dmitrii Podoprikhin, Timur Garipov, Dmitry Vetrov, and
  Andrew~Gordon Wilson.
\newblock Averaging weights leads to wider optima and better generalization.
\newblock In {\em 34th Conference on Uncertainty in Artificial Intelligence
  2018, UAI 2018}, pages 876--885. Association For Uncertainty in Artificial
  Intelligence (AUAI), 2018.

\bibitem{keskar2016large}
Nitish~Shirish Keskar, Dheevatsa Mudigere, Jorge Nocedal, Mikhail Smelyanskiy,
  and Ping Tak~Peter Tang.
\newblock On large-batch training for deep learning: Generalization gap and
  sharp minima.
\newblock {\em ICLR}, 2017.

\bibitem{kirkpatrick2017overcoming}
James Kirkpatrick, Razvan Pascanu, Neil Rabinowitz, Joel Veness, Guillaume
  Desjardins, Andrei~A Rusu, Kieran Milan, John Quan, Tiago Ramalho, Agnieszka
  Grabska-Barwinska, et~al.
\newblock Overcoming catastrophic forgetting in neural networks.
\newblock {\em PNAS}, page 201611835, 2017.

\bibitem{cifar}
Alex Krizhevsky, Geoffrey Hinton, et~al.
\newblock Learning multiple layers of features from tiny images.
\newblock 2009.

\bibitem{lange2020unsupervised}
Matthias~De Lange, Xu Jia, Sarah Parisot, Ales Leonardis, Gregory Slabaugh, and
  Tinne Tuytelaars.
\newblock Unsupervised model personalization while preserving privacy and
  scalability: An open problem.
\newblock In {\em Proceedings of the IEEE/CVF Conference on Computer Vision and
  Pattern Recognition}, pages 14463--14472, 2020.

\bibitem{mnist}
Yann LeCun, L{\'e}on Bottou, Yoshua Bengio, and Patrick Haffner.
\newblock Gradient-based learning applied to document recognition.
\newblock {\em Proceedings of the IEEE}, 86(11):2278--2324, 1998.

\bibitem{li2016learning}
Zhizhong Li and Derek Hoiem.
\newblock Learning without forgetting.
\newblock In {\em ECCV}, pages 614--629. Springer, 2016.

\bibitem{liu2018rotate}
Xialei Liu, Marc Masana, Luis Herranz, Joost Van~de Weijer, Antonio~M Lopez,
  and Andrew~D Bagdanov.
\newblock Rotate your networks: Better weight consolidation and less
  catastrophic forgetting.
\newblock In {\em 2018 24th International Conference on Pattern Recognition
  (ICPR)}, pages 2262--2268. IEEE, 2018.

\bibitem{lomonaco2021avalanche}
Vincenzo Lomonaco, Lorenzo Pellegrini, Andrea Cossu, Antonio Carta, Gabriele
  Graffieti, Tyler~L. Hayes, Matthias~De Lange, Marc Masana, Jary Pomponi, Gido
  van~de Ven, Martin Mundt, Qi She, Keiland Cooper, Jeremy Forest, Eden
  Belouadah, Simone Calderara, German~I. Parisi, Fabio Cuzzolin, Andreas
  Tolias, Simone Scardapane, Luca Antiga, Subutai Amhad, Adrian Popescu,
  Christopher Kanan, Joost van~de Weijer, Tinne Tuytelaars, Davide Bacciu, and
  Davide Maltoni.
\newblock Avalanche: an end-to-end library for continual learning, 2021.

\bibitem{lopez2017gradient}
David Lopez-Paz and Marc'Aurelio Ranzato.
\newblock Gradient episodic memory for continual learning.
\newblock In {\em Advances in neural information processing systems}, pages
  6467--6476, 2017.

\bibitem{Mallya2018}
Arun Mallya, Dillon Davis, and Svetlana Lazebnik.
\newblock Piggyback: Adapting a single network to multiple tasks by learning to
  mask weights.
\newblock In {\em ECCV}, pages 67--82, 2018.

\bibitem{Mallya2017}
Arun Mallya and Svetlana Lazebnik.
\newblock Packnet: Adding multiple tasks to a single network by iterative
  pruning.
\newblock In {\em CVPR}, pages 7765--7773, 2018.

\bibitem{mirzadeh2021linear}
Seyed~Iman Mirzadeh, Mehrdad Farajtabar, Dilan Gorur, Razvan Pascanu, and
  Hassan Ghasemzadeh.
\newblock Linear mode connectivity in multitask and continual learning.
\newblock In {\em International Conference on Learning Representations}, 2021.

\bibitem{mirzadeh2020understanding}
Seyed~Iman Mirzadeh, Mehrdad Farajtabar, Razvan Pascanu, and Hassan
  Ghasemzadeh.
\newblock Understanding the role of training regimes in continual learning.
\newblock {\em arXiv preprint arXiv:2006.06958}, 2020.

\bibitem{prabhu2020gdumb}
Ameya Prabhu, Philip~HS Torr, and Puneet~K Dokania.
\newblock Gdumb: A simple approach that questions our progress in continual
  learning.
\newblock In {\em European Conference on Computer Vision}, pages 524--540.
  Springer, 2020.

\bibitem{Rebuffi2017}
Sylvestre-Alvise Rebuffi, Alexander Kolesnikov, Georg Sperl, and Christoph~H
  Lampert.
\newblock icarl: Incremental classifier and representation learning.
\newblock In {\em CVPR}, pages 2001--2010, 2017.

\bibitem{russakovsky2015imagenet}
Olga Russakovsky, Jia Deng, Hao Su, Jonathan Krause, Sanjeev Satheesh, Sean Ma,
  Zhiheng Huang, Andrej Karpathy, Aditya Khosla, Michael Bernstein, et~al.
\newblock Imagenet large scale visual recognition challenge.
\newblock {\em IJCV}, 115(3):211--252, 2015.

\bibitem{rusu2016progressive}
Andrei~A Rusu, Neil~C Rabinowitz, Guillaume Desjardins, Hubert Soyer, James
  Kirkpatrick, Koray Kavukcuoglu, Razvan Pascanu, and Raia Hadsell.
\newblock Progressive neural networks.
\newblock {\em arXiv preprint arXiv:1606.04671}, 2016.

\bibitem{seff2017continual}
Ari Seff, Alex Beatson, Daniel Suo, and Han Liu.
\newblock Continual learning in generative adversarial nets.
\newblock {\em arXiv preprint arXiv:1705.08395}, 2017.

\bibitem{Serra2018}
Joan Serra, Didac Suris, Marius Miron, and Alexandros Karatzoglou.
\newblock Overcoming catastrophic forgetting with hard attention to the task.
\newblock In {\em International Conference on Machine Learning}, pages
  4548--4557. PMLR, 2018.

\bibitem{DGR}
Hanul Shin, Jung~Kwon Lee, Jaehong Kim, and Jiwon Kim.
\newblock Continual learning with deep generative replay.
\newblock In {\em NeurIPS}, pages 2994--3003, 2017.

\bibitem{shin2017continual}
Hanul Shin, Jung~Kwon Lee, Jaehong Kim, and Jiwon Kim.
\newblock Continual learning with deep generative replay.
\newblock In {\em Proceedings of the 31st International Conference on Neural
  Information Processing Systems}, pages 2994--3003, 2017.

\bibitem{silver2016mastering}
David Silver, Aja Huang, Chris~J Maddison, Arthur Guez, Laurent Sifre, George
  Van Den~Driessche, Julian Schrittwieser, Ioannis Antonoglou, Veda
  Panneershelvam, Marc Lanctot, et~al.
\newblock Mastering the game of go with deep neural networks and tree search.
\newblock {\em nature}, 529(7587):484--489, 2016.

\bibitem{silver2018general}
David Silver, Thomas Hubert, Julian Schrittwieser, Ioannis Antonoglou, Matthew
  Lai, Arthur Guez, Marc Lanctot, Laurent Sifre, Dharshan Kumaran, Thore
  Graepel, et~al.
\newblock A general reinforcement learning algorithm that masters chess, shogi,
  and go through self-play.
\newblock {\em Science}, 362(6419):1140--1144, 2018.

\bibitem{van2020brain}
Gido~M van~de Ven, Hava~T Siegelmann, and Andreas~S Tolias.
\newblock Brain-inspired replay for continual learning with artificial neural
  networks.
\newblock {\em Nature communications}, 11(1):1--14, 2020.

\bibitem{miniimagenet}
Oriol Vinyals, Charles Blundell, Timothy Lillicrap, Koray Kavukcuoglu, and Daan
  Wierstra.
\newblock Matching networks for one shot learning.
\newblock In {\em Proceedings of the 30th International Conference on Neural
  Information Processing Systems}, pages 3637--3645, 2016.

\bibitem{zenke2017continual}
Friedemann Zenke, Ben Poole, and Surya Ganguli.
\newblock Continual learning through synaptic intelligence.
\newblock In {\em ICML}, pages 3987--3995. JMLR. org, 2017.

\bibitem{zhang2016understanding}
Chiyuan Zhang, Samy Bengio, Moritz Hardt, Benjamin Recht, and Oriol Vinyals.
\newblock Understanding deep learning requires rethinking generalization.
\newblock {\em arXiv preprint arXiv:1611.03530}, 2016.

\bibitem{hsu2018re}
Yen-Chang Hsu, Yen-Cheng Liu, Anita Ramasamy, and Zsolt Kira.
\newblock Re-evaluating continual learning scenarios: A categorization and case
  for strong baselines.
\newblock {\em arXiv preprint arXiv:1810.12488}, 2018.

\bibitem{van2019three}
Gido~M van~de Ven and Andreas~S Tolias.
\newblock Three scenarios for continual learning.
\newblock {\em arXiv preprint arXiv:1904.07734}, 2019.

\end{thebibliography}

\appendix
\addcontentsline{toc}{section}{Appendices}
\section*{Appendix}

The supplementary materials include:
\begin{itemize}[noitemsep,nolistsep]
    \item \textbf{Appendix A}: Additional experimental results for MNIST and CIFAR10.
    \item \textbf{Appendix B}: Proof of arguments in GEM discussion.
    \item \textbf{Appendix C}: Comprehensive reproducibility details.
\end{itemize}

\section{Additional results}
Due to space constraints in the main paper, in this section we report the additional results.

\begin{figure*}[!ht]
\centering
\includegraphics[clip,trim={0cm 0.5cm 0cm 0cm},width=.7\linewidth]{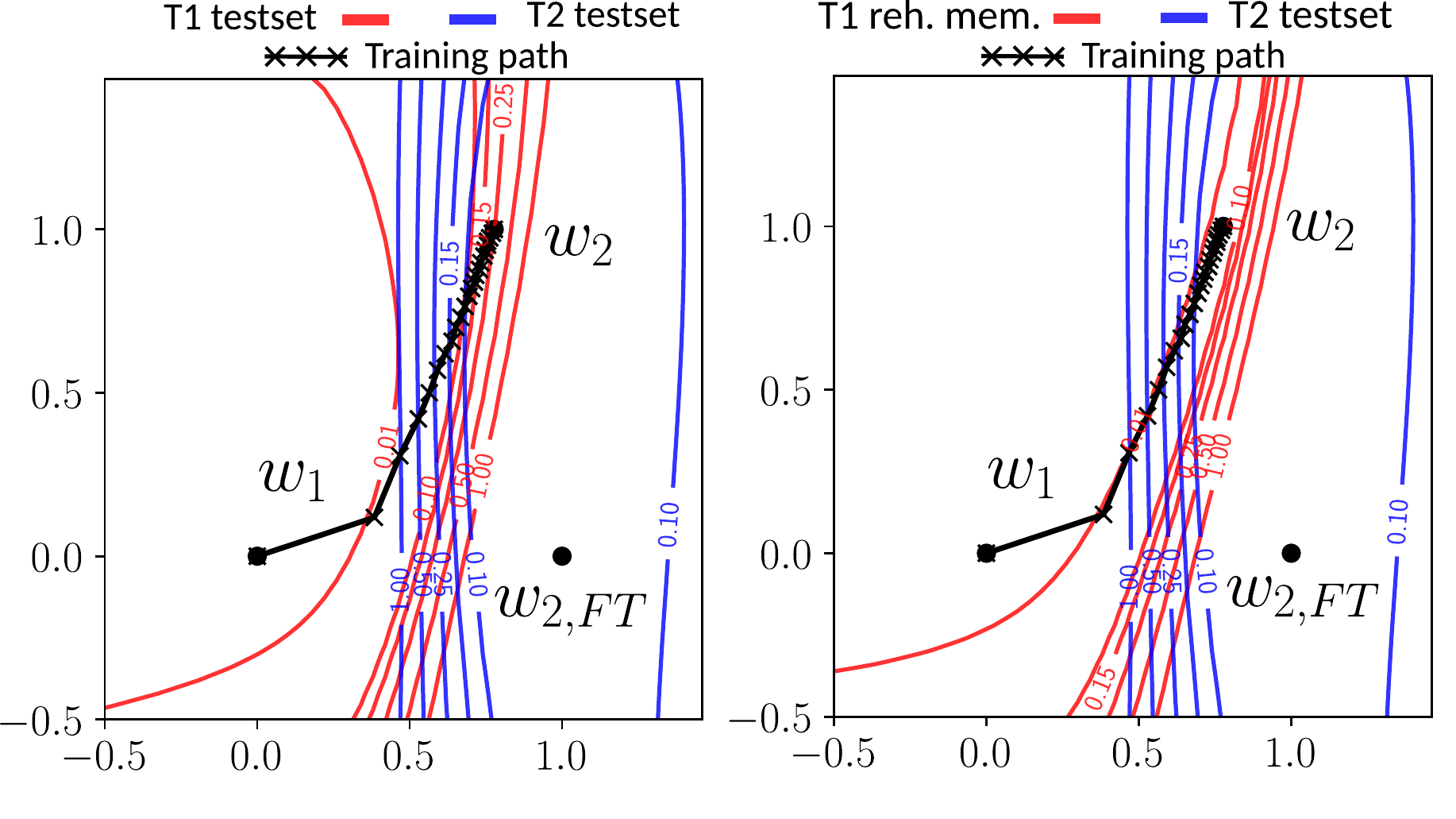}
   \caption{Projection of MNIST learning trajectories in parameter space on the plane defined by $w_1$, $w_2$ and $w_{2,FT}$.
   For the same $T2$ test loss (blue), the loss for $T1$ (red) is calculated in two different ways.
   \emph{left}: $T1$ loss for the vast test set. \emph{right}: $T1$ loss for the limited rehearsal memory.}
   \label{fig:contour_plots_mnist}
\end{figure*}

\boldspacepar{Learning trajectory MNIST.}
Figure~\ref{fig:contour_plots_mnist} illustrates the learning trajectory projection in parameter space for MNIST. The CIFAR10 and Mini-Imagenet results are reported in the main paper, for which the findings extend to this MNIST sequence as well.

\begin{figure*}[!ht]
    \centering
    \begin{subfigure}{1\linewidth}
    \caption*{MNIST} \vspace{-0.3cm}
    \includegraphics[clip,trim={0cm 0cm 0cm 0cm},width=\linewidth]{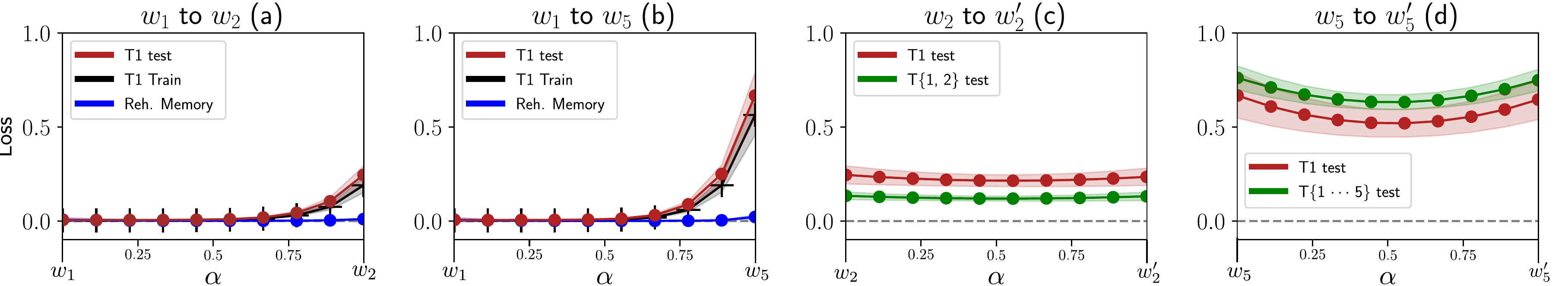}
    \end{subfigure} \\
    \begin{subfigure}{1\linewidth}
        \vspace{0.3cm}\caption*{CIFAR10}\vspace{-0.3cm}
    \includegraphics[clip,trim={0cm 0cm 0cm 0cm},width=\linewidth]{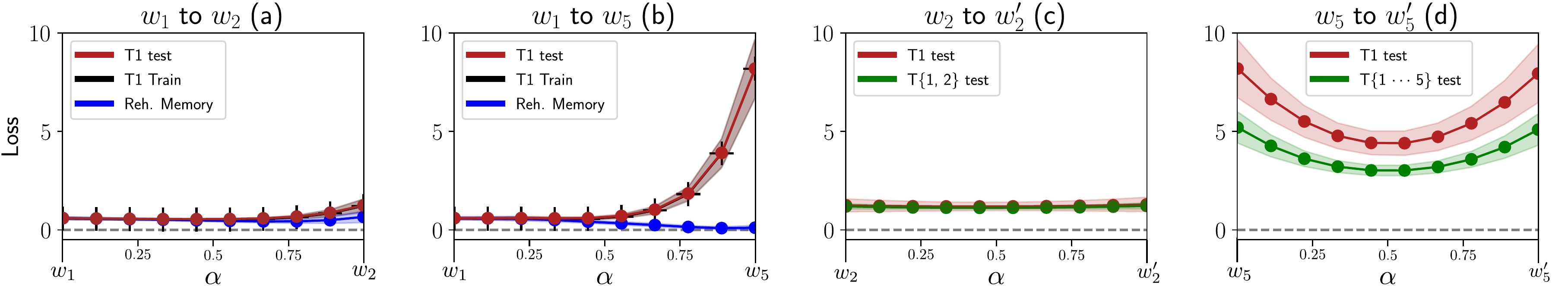}
    \end{subfigure} 
    \caption{Avg.\ loss and standard deviation on linear paths between the models used in the empirical evidence of hypotheses one and two. Training is performed on MNIST (\emph{top}) and CIFAR10 (\emph{bottom}) and sampled 100 times for different model initializations and memory populations.
    A path from $w_i$ to $w_j$ is calculated as $(1-\alpha)w_i + \alpha w_j$. 
    \emph{(a) and (b)}: %
    Loss on the linear path between the model after learning $T1$ ($w_1$) and the model after learning with rehearsal on $T2$ and $T5$ respectively. 
    \emph{(c) and (d)}: %
    Loss of the path between two models learned with different memory populations, after $T2$ and $T5$ respectively. Red is the loss on the T1 testset, green the average loss of all tasks up to T2 and T5 respectively.
     Results contained no outliers with a higher loss on the path compared to the loss of the models.
    }
    \label{fig:paths:MNIST_CIFAR}
\end{figure*}

\boldspacepar{Loss of linear  interpolations MNIST and CIFAR10.}
Figure~\ref{fig:paths:MNIST_CIFAR} reports the loss for linear interpolations in parameter space for MNIST and CIFAR10. Results for Mini-Imagenet are reported in the main paper.
Notably, CIFAR10 $w_1$ to $w_2$ does not overfit significantly on the rehearsal memory. However, training is only done for one epoch per task and after training three more tasks with the rehearsal memory, $w_5$ reports near zero loss for the rehearsal memory, indicating overfitting. This shows that rehearsal may overfit more on the rehearsal memory as the training sequence length increases, either by more tasks (e.g. CIFAR10) or more epochs per task (e.g. Mini-Imagenet with 10 epochs per task).

\boldspacepar{MNIST high-loss ridge aversion.} 
We provide additional experiments for other commonly used rehearsal memory sizes ($0.2$k and $2$k) in the MNIST setup~\cite{de2020continual,hsu2018re,van2019three}.

\begin{table}[!ht]
\centering
\begin{tabular}{@{}lll@{}}
\toprule
\textbf{\textbf{Experiment}} & \multicolumn{2}{c}{\textbf{Rehearsal memory size $ \bf |\mathcal{M}|$}} \\ \midrule
                             & \textit{200}                   & \textit{2000}                \\
\textbf{MNIST}               &                                &                              \\
\textit{ER}                  & $81.8 \pm 0.7$                 & $91.8 \pm 0.4$               \\
\textit{ER-step}             & $87.6 \pm 1.1$ (n=20)          & $92.6 \pm 0.3$ (n=5)         \\ \bottomrule
\end{tabular}%
\caption{Additional MNIST avg.\ accuracy results for memory sizes $200$ and $2000$, comparing ER and ER-step for the high-loss ridge aversion experiment with $n$ the number of steps.}
\label{tab:my-table}
\end{table}

\section{Loss constraints: GEM vs.\ Rehearsal}
In GEM \cite{lopez2017gradient}, the updates of the model are restricted to the directions where the loss on the memory samples decreases or remains equal. This is imposed by requiring $g_n \cdot g_i \geq 0, \forall i$ with $g_n$ the gradient on the new batch and $g_i$ on sample $i$ in the rehearsal memory. 
This is a first-order approximation, hence it is only exact where the loss surface is linear.

In contrast, in rehearsal an increase of the loss on the memory samples is allowed, as long as it is smaller or equal than the decrease in loss on the new batch. We proof this in the following.

A model update with SGD is calculated as:
\begin{equation}
    w' \leftarrow w - \alpha g
\end{equation}
with gradient $g$ and learning rate $\alpha$. If we assume the first order approximation to hold in an $\alpha$-region around $w$, then because the negative gradient is either zero or points in a direction with decreasing loss, it follows that
\begin{align}
    \mathcal{L}(w') &\leq \mathcal{L}(w) \nonumber \ , \\ 
    \mathcal{L}_m(w') + \mathcal{L}_n(w') &\leq \mathcal{L}_m(w) + \mathcal{L}_n(w) \nonumber  \ , \\
    \mathcal{L}_m(w') - \mathcal{L}_m(w) &\leq \mathcal{L}_n(w) - \mathcal{L}_n(w') \ , 
    \label{eq:loss_increase}
\end{align}
with $\mathcal{L}$ the average of the loss on the memory $\mathcal{L}_m$ and the loss of the new batch $\mathcal{L}_n$. Therefore, based on the same first order approximation as in \cite{lopez2017gradient}, Eq.~\ref{eq:loss_increase} shows that rehearsal only allows increases in loss on the memory as large as the decrease in loss on the new batch.

\section{Reproducibility details}
This section provides all the details to maintain reproducibility of our experiments. Furthermore, our codebase provides the original implementation in Pytorch to reproduce our results.

\subsection{Empirical evidence Hypotheses 1 and 2}
\label{apdx:exp:hyp12}
\boldspacepar{MNIST} is trained with a two-layer MLP, with each layer 400 ReLU nodes. 
Optimization of the model uses vanilla stochastic gradient descent (SGD), 
with a constant learning rate of $0.01$. Each update is performed on a batch of 10 new and 10 memory samples.
The rehearsal memory has a fixed capacity of 50 samples per task. This fixed capacity is allocated before training to enable analyzing overfitting for static task-specific rehearsal memory's.
Online training is performed as each sample is only seen once during training, except for the memory samples. The MNIST split results in $T1$ containing 0's and 1's and $T2$ containing 2's and 3's. 

\boldspacepar{CIFAR10} training details are equal to those of MNIST, except for the model and the memory capacity. The model used is the reduced Resnet18, introduced by by Lopez-Paz et al.\ \cite{lopez2017gradient}. $T1$ of CIFAR10  consists of planes and cars and $T2$ contains birds and cats, following the standard split. The memory capacity is 100 samples per task.

\boldspacepar{Mini-Imagenet} training details are equal to those of CIFAR10, except for training 10 epochs per task rather than training online. For Mini-Imagenet there is no standard split and the categories were assigned randomly to a task, but remained the same in all experiments. As in CIFAR10, the memory capacity is 100 samples per task.

\subsection{High-loss ridge aversion}
This section details the learning details for the high-loss ridge aversion experiments. 
All benchmarks use a gridsearch for the number of steps $n \in \left\{0,1,2,3,4,5,10,20,50\right\}$, with $n=0$ the Experience Replay (ER) baseline. We report the best results from this gridsearch for \emph{ER-step} with $n>0$, following the procedure in \cite{lopez2017gradient}. All results are obtained with $10$ epochs per task.
In contrast to the hypothesis experiments discussed in Appendix~\ref{apdx:exp:hyp12},
this experiment allows for a dynamically subdivided rehearsal memory instead of a fixed allocation over all tasks. We use this memory policy in this experiment as it is commonly used in literature and allows exploiting the full memory capacity~\cite{Rebuffi2017,de2020continual,9349197}.

\boldspacepar{MNIST} has the same setup as in Appendix~\ref{apdx:exp:hyp12}, except with 10 epochs per task and learning rate $0.001$ to allow smaller steps when training only on the rehearsal memory.

\boldspacepar{CIFAR10 and Mini-Imagenet}
follow the reduced Resnet18 setup with Stable-SGD~\cite{mirzadeh2020understanding} for CIFAR100 in~\cite{mirzadeh2021linear}. That is, Stable-SGD is used to obtain wider minima, with initial learning rate $0.1$, decayed per task with factor $0.8$ and with momentum $0.8$. The fixed dropout rate $0.1$ is obtained from gridsearch in values $\left[0.1,0.25\right]$.

\subsection{Projection plots}
The loss contour plots in the parameter space  as in Figure~\ref{fig:contour_plots_mnist} are inspired by recent work \cite{mirzadeh2021linear}. They show a hyperplane in the parameter space, defined by three points $w_1$, $w_2$ and $w_3$. Orthogonalizing $w_2 - w_1$ and $w_3 - w_1$ gives a two dimensional coordinate system with base vectors $u$ and $v$. The value at point $(x, y)$ is then calculated as the loss of a model with parameters $w_1 + u \cdot x + v \cdot y$. For more details, we refer to our code or the appendix in \cite{mirzadeh2021linear}.
The training trajectories shown in these figures are from a single run and are the projections of the points in the parameter space to this hyperplane. The projection of $w'$ is calculated as $u \cdot (w' - w_1)$ and $v \cdot (w' - w_1)$. The indicated points are each 50, 50 and 100 steps apart for respectively MNIST, CIFAR10 and Mini-Imagenet.

\end{document}